\documentclass[letterpaper]{article}
\usepackage[preprint]{aaai2027}
\ifXeTeX
\usepackage{fontspec}
\fi
\usepackage[hyphens]{url}
\usepackage{natbib}
\usepackage{amsmath,amssymb,booktabs,array,microtype}
\usepackage{graphicx,xcolor,tikz}
\usetikzlibrary{arrows.meta,positioning,shapes.geometric,calc}
\usepackage{listings}
\definecolor{policyblue}{RGB}{47,88,135}
\definecolor{evalorange}{RGB}{173,171,132}
\definecolor{memteal}{RGB}{165,144,131}
\definecolor{agentplum}{RGB}{112,66,138}
\newcommand{\reach}{\operatorname{reach}}
\newcommand{\acquire}{\operatorname{acquire}}
\newcommand{\maintain}{\operatorname{maintain}}
\newcommand{\mix}{\operatorname{mix}}
\newcommand{\E}{\mathbb{E}}
\newcommand{\ind}{\mathbf{1}}
\newcommand{\FCFinalArrival}{50.49}
\newcommand{\FCWholeArrival}{0.29}

\newcommand{\FCSharedSuccess}{71.30}
\newcommand{\FCGuardsSuccess}{36.20}

\title{Continuation-Guided LLM Evolution of Program Policies \\and Goal Compilers in Craftax }
\author{Víctor Gallego}
\affiliations{Komorebi AI Technologies}
\begin{document}
\maketitle

\begin{abstract}
Coding agents can synthesize and optimize code-based policies from simulator feedback, allowing the resulting program to act with zero inference model calls. But for long horizon tasks, two challenges arise: local improvements can hurt final success and switching to a new objective requires another expensive synthesis run. We propose a framework that first develops competence by refining the policy through a feedback cascade, moving from episode returns to hierarchical goals and belief updates which can be scored densely from simulator. In the Craftax game, a partially observed and procedurally generated environment, refinements via episode level feedback stall at <1\% final-level arrival, whereas with the feedback cascade the optimized policy reaches the final stage 50\% of time (28\% final boss defeats). Furthermore, to make this competence reusable towards new goals, we factor the previous controller into an environment interpreter $I$ (models, memory, skills) and a rule program $p_0$, and we show that coding agents can optimize a reusable goal compiler $C_{\theta}$ that maps structured tasks $g$ into new rule programs $p_g$. Across five independent compiler searches trained only on short tasks, the resulting compilers successfully transfer zero-shot to longer tasks without any task-specific tuning nor inference-time LLM calls: continuation-scored feedback and a fixed machine with an editable program together turn a single-task competent policy into a reusable one. \\

Code at: \url{https://github.com/vicgalle/craftax-program-policies}.
\end{abstract}

\newlength{\savedparskip}
\setlength{\savedparskip}{\parskip}
\setlength{\parskip}{2pt plus .5pt minus .5pt}
\section{Introduction}

Language models can produce executable control programs and refine them using simulation feedback \citep{codepolicies,nesyprog}. During development, a coding agent edits a program (with explicit state, rules, and memory); at deployment, this artifact is used as a policy, without an LLM call at each action. This is an example of neuro-symbolic division of labor, separating expensive search from cheap and auditable execution. The feedback (i.e., what the model sees between refinements) determines the space of feasible policies \citep{gallego}. In long-horizon tasks, where a decision maker coordinates behaviors over tens of thousands of actions, this paradigm raises two related questions: what feedback makes code-based policy search effective, and which artifacts should that search leave behind?

Consider Craftax \citep{craftax}, a procedurally generated survival game in which an agent has 100,000 actions to gather resources, craft tools and armor, manage physical state, combat enemies, and descend through eight increasingly difficult levels. For an early policy, reaching the last level happens in a negligible fraction of episodes and depends on decisions made thousands of steps earlier; and using scalar episode rewards quickly confounds the relevant failure. Decomposing the task into stages or subgoals introduces a subtle trap: \emph{we shouldn't use a subgoal's local success as a source of value, instead, use the continuation it leaves behind}. For instance, an agent exploring the Ice Realm (floor 7) may interrupt exploration to fetch food on the previous level and return, but this strategy helps only if the agent survives the trip with enough health and time left to finish the rest of the journey. In the experiments, we observe that across coding agent's code updates, an edit's effect on reaching the next level vs the same edit's effect on reaching the final level are negatively correlated. Hence, the useful statistic to accept a policy code update need not coincide with a subgoal's completion condition.

\begin{figure*}[t]
\centering
\resizebox{\textwidth}{!}{\input{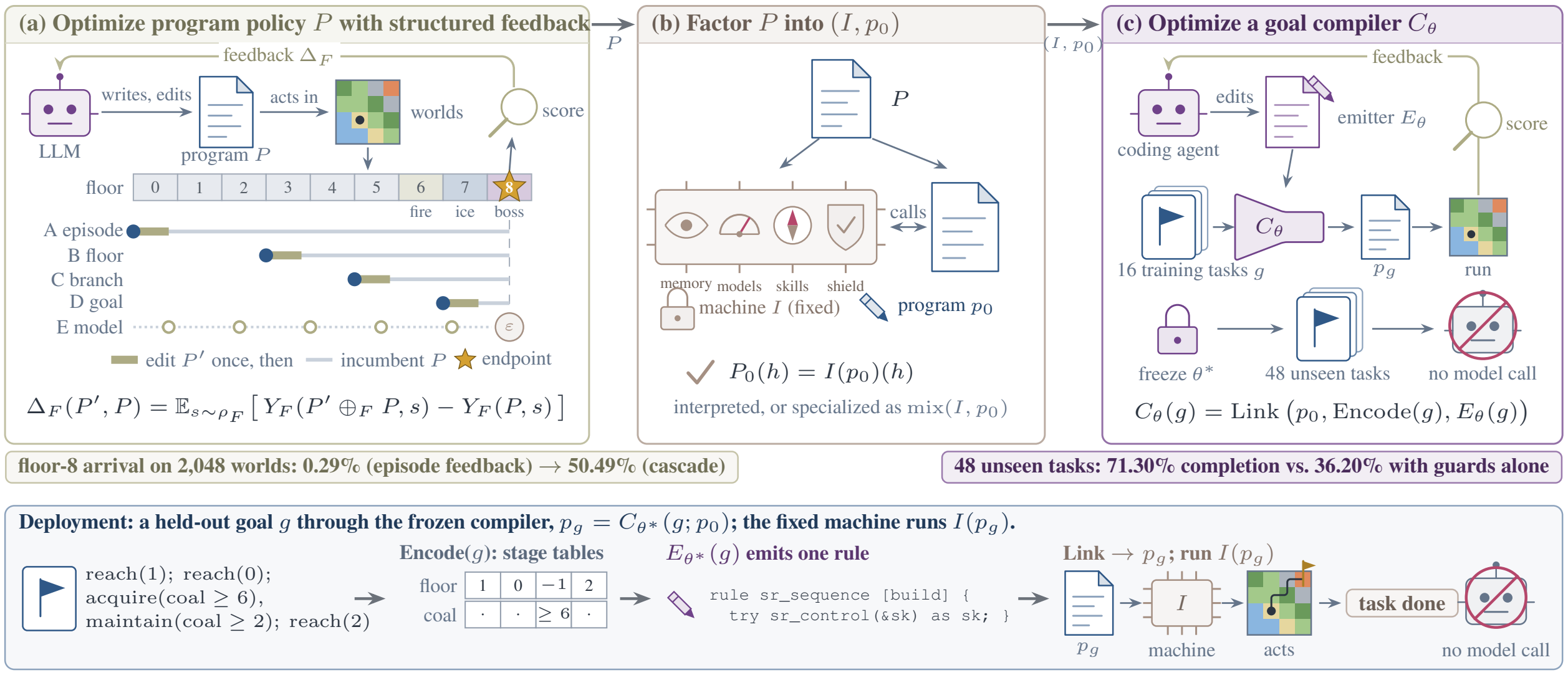}}
\caption{From policy development to task reuse. (a) Coding agents refine a program policy $P$ through a feedback cascade (A-E). (b) The resulting policy is factored into a fixed machine $I$ and a rule program $p_0$, with performance guarantees. (c) Agents optimize a goal compiler $C_\theta$ on a small set of short tasks; then at deployment (bottom), the frozen compiler translates unseen, longer tasks $g$ into new policies $p_g$ without LLM calls at inference time.}
\label{fig:overview}
\end{figure*}

Our framework operates in two stages. Firstly, it proposes a feedback cascade so a coding agent can progressively optimize a code-based policy towards the objective. Over a test set of 2048 game worlds, gross episode-level feedback stalled at 0.29\% final floor arrivals, whereas the cascade pushed the arrival rate to 50.49\%. Further control experiments confirm that coding agents benefit from the proposed harness.
Secondly, an already optimized policy creates an opportunity after refinement: its world models, spatial memory and skills encode useful domain knowledge, but this is interleaved with procedural information specialized to the original objective. Our second stage makes that boundary explicit and turns a single-task policy into a reusable target for code generation of new policies for alternative objectives. Figure~\ref{fig:overview} depicts the framework.

 Our paper makes three contributions: (i) a framework for coding agents evolving code-based policies that are exposed to a cascade of progressively structured feedback, (ii) a policy optimization study with strict cohort evaluation and an audit of misleading local feedback, (iii)
a compiler search experiment testing transfer to unseen task
compositions, including longer sequences, without model calls at inference.

\section{Framework}
\label{sec:framework}
The framework has two stages: improve a policy using structured execution feedback towards a pre-declared objective, then turn its knowledge into a compiler that produces new programs for alternative objectives. We first define the policy and goal interfaces used by both stages.

\subsection{Policies and Goals}
\label{sec:setup}

Craftax Full\footnote{We use the C language version of the environment available at\\ \url{https://github.com/Infatoshi/craftax.cu}.} has nine floors, from the overworld (0) to the Graveyard (8), where the final boss (Necromancer) is to be defeated. Using RL terminology, at every step the observation is a local symbolic map with items, creatures, inventory and intrinsics. To play competently at the game, the agent has to learn its own inner maps of visited floors, ladder locations and resource state.
The policy never receives hidden state from the engine at execution. Coding agents can read simulator source code during
development (so they are aware of the rules of the game), and can restore complete simulator snapshots to restart an agent from a particular state. The policy is a C program, rather than a neural policy trained by gradient descent.

Let $X_t$ be the hidden game state, $O_t$ the public observation and $Z_t$
the policy's persistent memory. At every step $t$, the policy $P$ produces an action $A_t$ and a memory
update, potentially controlled by an active internal goal $u_t$:
\begin{equation}
 A_t=\pi_{u_t}(Z_t),\qquad Z_{t+1}=U(Z_t,A_t,O_{t+1}).
 \label{eq:memory}
\end{equation}
$Z_t$ includes the current observation and goal context. The policy can have components modeling aspects from the game, i.e.
computing beliefs $\hat y_t=m(Z_t)$ about true but unobserved quantities $y_t=\phi(X_t)$ such as
incoming damage, enemy cooldowns or resource depletion. 

For an external objective $g$, let's write $J_g(P)=\E[Y_g(P)]$, where $Y_g$ is a binary terminal success indicator. Long tasks often pass through ordered milestones $0,\dots,K$, where milestone $k+1$ can only be reached after milestone $k$. If $E_k$ denotes reaching milestone $k$ within the horizon, with $E_0$ certain, then the objective
$g=$ ``reach milestone $K$'' can be factorized as
\begin{equation}
 J_g(P)=\Pr_P(E_K)=\prod_{k=0}^{K-1}\Pr_P(E_{k+1}\mid E_k).
 \label{eq:funnel}
\end{equation}
In Craftax the milestones are the floors (levels) with $K=8$ and the horizon is 100,000 steps. Making this factorization explicit helps in locating bottlenecks. For instance, upstream changes alter downstream equipment, health, memory and time, while recovery trips create recurring behavior across floors. 

\paragraph{Internal and external goals.}
Because the word "goal" appears in both stages of our workflow, we distinguish internal goals ($u$) from external ones ($g$).

\emph{Internal goals} are persistent subroutines inside the policy that organize behavior and code during policy optimization (our stage 1) and inference. They are related to RL options and controlled hierarchical hidden Markov models \citep{options,murphy}, as each internal goal $u$ has symbolic entry, completion and abandonment predicates and executes on a call stack. For example, a parent goal exploring the Ice Realm (floor 7) remains active on the stack while a (child) subgoal starts, producing an excursion going back to Fire Realm (floor 6) for resources such as food. At each step, the controller checks if $u$ continues, completes or abandons (returning control to parent goal in the later two cases). These invocations define the start and end states for continuation-based feedback.

\emph{External goals} $g$ are multi-stage objectives supplied at inference time. For example: reach floor 1, return to floor 0, acquire six coal, and reach floor 2 while maintaining two coal. And this is regardless of which internal goals the policy controller calls along the way.

\subsection{Feedback Cascade}
\label{sec:feedback}
In our code-based policy search setup, each refinement round starts from an incumbent policy $P$. A coding agent inspects source code, receives feedback, proposes a candidate edit $P'$, and tests it in simulation. An evaluator estimates this statistic, which measures the difference:
\begin{equation}
 \Delta_F(P',P)=\E_{s\sim\rho_F}\!\left[
 Y_F(P'\oplus_F P,s)-Y_F(P,s)\right].
 \label{eq:feedback}
\end{equation}
In every comparison, the candidate run and the incumbent run start from the exact same state $s$ (either a fresh world at reset, or a restored simulator snapshot) and share the same initial random seed. Pairing on $(s, \mbox{seed})$ cancels world-to-world variance, though trajectories and in-game random draws diverge once $P'$ selects a different action.

Across the cascade, the regimes progressively refine
where comparisons start ($\rho_F$), how long the candidate acts ($P'\oplus_F P$) and which outcome is measured ($Y_F$). Table~\ref{tab:regimes} lists the five instantiations.

\begin{table*}[t]
 \centering
 \footnotesize
 \begin{tabular}{@{}llp{3.6cm}p{3.5cm}p{5.6cm}@{}}
 \toprule
 & Rounds & Start states $\rho_F$ & Acting program $P'\oplus_F P$ & Statistic $Y_F$ \\
 \midrule
 A & 1--10 & fresh worlds & $P'$ throughout & return (rounds 1--4); reaching the frontier floor (5--10) \\
 B & 11--13 & fresh worlds in which $P$ reaches floor $k$ & $P'$ throughout & reaching floor $k{+}1$ \\
 C & 14 & saved floor-milestone states & $P'$ from $s$ on & reaching floor 8 \\
 D & 15--16, ext. & saved floor and goal entries; fresh worlds & $P'$ from $s$ on (repeated use); $g'$ once, then $P$; $P'$ from reset & reaching floor 8 \\
 E & ext. & every step of incumbent runs & $P$ (the candidate's beliefs computed, not read) & negative belief error against $X_t$ \\
 \bottomrule
 \end{tabular}
 \caption{The five feedback regimes as instances of
 Equation~\eqref{eq:feedback}. All comparisons pair candidate and incumbent on
 the same worlds or saved states with the same random stream. Ext.: the runs after round~16
 (Appendix~\ref{app:lineage}).}
 \label{tab:regimes}
\end{table*}

\paragraph{A: Episode outcomes.} Both $P'$ and $P$ run from fresh world resets to the end of the episode.  The score can be the environment returns or the rate of arrivals to the final level. While simple, this provides no signal for late-game code that is rarely reached.
\paragraph{B: Per-milestone pass rates.} This assigns coding sub-agents to individual milestones from Eq. (\ref{eq:funnel}). If a code edit shouldn't fire before the $k-$th milestone (floor in Craftax), the harness verifies that $P'$ and $P$ produce identical actions up to that point. It also measures arrival at milestone $k+1$ (using separate seeds for confirmation and integration). However, optimizing $k \rightarrow k+1$ is myopic wrt the final milestone.
\paragraph{C: Save-state branching.} Instead of re-runs from reset, the harness captures snapshots (engine state, RNG key, policy memory) when the incumbent reaches new milestones. Then, both $P$ and $P'$ resume from the saved snapshot and run through episode end (with score the final arrival, $Y_F = \ind_{E_8}$).
\paragraph{D: Persistent goals.} Milestones still can conflate behaviors that extend across several of them, such as leaving floor 7 to fetch food on floor 6. Within this regime, the coding agent is prompted to use internal goals ($u$) as the unit of code ownership and evaluation, capturing snapshots at goal entry and comparing candidate $P'$ from there on eventual final milestone arrival.
\paragraph{E: Model audits.} Finally, for edits that update internal beliefs $\hat y_t=m(Z_t)$ without changing actions, $P'$ is evaluated along $P$ by measuring belief accuracy against the simulator true state at every step.

Because saved snapshots $\rho_F$ sample from the incumbent's state distribution rather than the candidate's, integrated edits are always verified from fresh resets and snapshots are recaptured after upstream updates.

\subsection{Decision and Model Refinements}
A code edit may alter \emph{decisions} (actions the policy emits) or \emph{models} (how internal beliefs $\hat y_t=m(Z_t)$ are computed). Because decision edits change the trajectory but model edits do not, we use separate approaches to evaluate them.

\paragraph{Continuation value.} Let $h$ be the public interaction history and active goal stack at a goal entry, and $b$ the remaining step budget. Write $H_{P,g}(h,b)$ for the incumbent $P$'s probability of eventually completing $g$. Replacing one goal invocation with a candidate $u$ that runs for $\tau$ steps yields rollout value
\begin{equation}
 Q_{P,g}(h,u,b)=\E\!\left[
 \ind_S+\ind_C H_{P,g}(h',b-\tau)\mid h,u,b\right],
 \label{eq:continuation}
\end{equation}
where $S$ is terminal success during $u$, and $C$ is an alive return (either completion or abandonment) leaving history $h'$ and budget $b-\tau$, after which $P$ resumes control. Equation (\ref{eq:continuation}) adapts rollout-based policy improvement over options \citep{tesauro,options} to programmatic subroutines. We estimate $Q_{P,g} - H_{P,g}$ directly from paired continuations from saved snapshots rather than fitting a value function. And crucially, Eq. (\ref{eq:continuation}) explains why subgoal completion and value diverge: a child that completes with probability 0.9 but leaves the agent depleted ($H_{P,g}=0.2$, overall value 0.18) is worse than one with completion 0.8 and leaving the agent well supplied  ($H_{P,g}=0.5$, overall 0.40).

\paragraph{Evaluation scopes.} Since each scope catches a different failure mode, the harness evaluates decision edits at three scopes (Fig. 1a): (i) one invocation from a saved goal entry (handing control back to $P$ at return) as a fast check; (ii) repeated use from parent entries (enabling $u$ whenever eligible) to test cumulative cost and repeated calls; and (iii) composition from reset on fresh worlds to verify that the merged edit works well under the induced state distribution.

\paragraph{Simulator feedback for model edits.} In the case of model edits, for a belief module $m(Z_t)$ estimating hidden state $\phi(X_t)$, the harness measures step by step prediction error for some regression loss $\ell$:
\begin{equation}
 \varepsilon(m)=\E_t\big[\ell(m(Z_t),\phi(X_t))\big].
 \label{eq:model}
\end{equation}
When a candidate model $m'$ is computed alongside the incumbent policy, but its updated beliefs are not used by the former, this model edit doesn't change the agent trajectory: using paired random seeds, candidate and incumbent policies take identical actions and visit same states. Thus $\Delta\varepsilon = \varepsilon(m') - \varepsilon(m)$ can be evaluated with zero outcome noise. Value enters then only when a policy is finally wired to read the new belief. Then the edit becomes a decision edit and must pass the continuation value test.

\paragraph{Counterfactual interventions.} To locate bottlenecks before editing code, the harness can also inject counterfactuals (e.g., free armor or refilled intrinsics at entry states) or replace beliefs $m(Z_t)$ with ground truth $\phi(X_t)$. These probes are only used to estimate potential ceilings to guide the search process.

\subsection{Policy Factorization}
\label{sec:factorization}
The optimized policy $P^*$ from the previous stage is a monolithic C program, in which general domain competence is mixed with a strategy that is specialized for reaching the final milestone (floor 8 in Craftax). In order to reuse that competence for new objectives, we factor $P^*$ into a fixed domain machine $I$ and an editable rule program $p_0$, which are both code artifacts (Figure~\ref{fig:overview}b).

The machine $I$ contains what is needed to survive and act in Craftax: observation processing, persistent spatial memory, world and resource models, navigation and crafting skills, etc. The program $p_0$ says which strategy to follow, and it is written in a teleo-reactive rule language \citep{nilsson}. At each step the rules are checked in priority order (survival and combat rules at the top, strategic progression rules below them) until the first rule that matches emits an action. A rule can read the public observation, query the beliefs of the machine, update the memory it is allowed to write, and call the skills of the machine. In the other direction, the machine can ask $p_0$ for strategy parameters such as reserve thresholds. Note that neither $I$ nor $p_0$ has access to the hidden simulator state.

A rule program $p$ can be interpreted by $I$ at run time, or it can be compiled into a specialized C++ binary by partial evaluation \citep{futamura}:
\begin{equation}
 P_0(h)=I(p_0)(h),\qquad t_0=\mix(I,p_0).
 \label{eq:specialize}
\end{equation}
We also require a strict empirical parity check. The original C policy $P^*$, the interpreted program $I(p_0)$ and the specialized binary $t_0$ must pick the same action at every step, over a large set of seeds (Section~\ref{sec:parity}). If this holds, the domain dynamics and primitives are kept fixed inside $I$, while the high-level strategy in $p_0$ is now a clean target for code generation, and this is what lets us reuse it for new goals.

\subsection{Goal Compilation}

Once $(I, p_0)$ is fixed, adapting the agent to a new external task $g$ becomes a problem of code generation. Instead of calling a coding agent to write a new policy every time a new task arrives, we use coding agents only during a development phase, in order to optimize a reusable \emph{goal compiler} $C_\theta$ (Figure~\ref{fig:overview}c):
\begin{equation}
\begin{aligned}
C_\theta(g; p_0) &= \mathrm{Link}\!\left(p_0, \mathrm{Encode}(g), E_\theta(g)\right), \\
p_g &= C_\theta(g; p_0), \qquad a_t = I(p_g)(h_t).
\end{aligned}
\end{equation}
As in stage 1, the coding agent optimizes a program, but this program now plays the role of a compiler instead of a specialized policy. This means that $\theta$ is executable Python code instead of a set of network weights. Given a task specification $g$, a fixed translator $\mathrm{Encode}(g)$ parses the task into constant stage tables, and the learned emitter $E_\theta(g)$ writes the helper definitions for that task and a single control rule. The linker $\mathrm{Link}$ puts this rule into the base program $p_0$ at a fixed priority slot \emph{below} the inherited survival and safety rules, and it disables the original floor-8 descent controller of $p_0$. Using what $I$ already knows, the emitter can adapt resource gathering, safety buffers and floor routing to the task.

\paragraph{Task language and verification.}
In our experiments, a task is an ordered sequence $g = (s_1, \dots, s_K)$ of at most eight stages. A stage can be a navigation milestone $\mathtt{reach}(\ell)$, where the agent has to visit floor $\ell$, or a resource milestone $\mathtt{acquire}(v)$, where it has to hold an inventory stock $v$ (and items that it already holds also count). A resource stage can also have a persistent maintenance constraint $\mathtt{maintain}(r)$, with $r \le v$. The agent has to complete the stages in order and without dying. When a stage with $\mathtt{maintain}(r)$ is completed, the reserve $r$ is added coordinatewise to the active reserves, and these have to be kept until the end of the task. At run time this is enforced with a hard action filter, which does not allow spending below the active reserves. To check progress we use an independent monitor that looks directly at the simulator state and does not trust what the controller declares.

\paragraph{A concrete compilation example.}
As an example, take a held-out task $g$ with four stages: \emph{reach floor 1, return to the surface (floor 0), acquire six coal while maintaining at least two coal, and then reach floor 2} (Figure~\ref{fig:overview}, bottom; full source in Appendix~\ref{app:example}). From it, the fixed translation $\mathrm{Encode}(g)$ builds the target floor array $[1, 0, -1, 2]$ ($-1$ marks the acquisition stage) and the coal stock and hold arrays $[0, 0, 6, 0]$ and $[0, 0, 2, 0]$. One of our selected emitters looks at this structure and decides that $g$ needs preparation for several floors. It then adjusts the equipment buffers and emits a clause for upward travel, which the return stage uses. At deployment, all we need to compile a new task is to run the frozen Python function $C_{\theta^*}$ and execute $I(p_g)$, without LLM calls and without tuning with rollouts.

\section{Experiments}
\label{sec:experiments}
In this section we look at three questions: how the policy is optimized, whether it can be factorized as described in Section \ref{sec:factorization}, and how well it transfers to new goal programs. 

\subsection{Policy Optimization}
\label{sec:development}

We evolved a single C policy across 25 checkpoints using Codex (\texttt{gpt-6-astra}) and Claude Code (\texttt{claude-opus-5-5}) agents (full details in Appendix \ref{app:protocols}, example prompts in Appendix \ref{app:prompts}). The policy was optimized through the five feedback regimes: whole-episode outcomes (rounds 1-10), per-floor pass rates (rounds 11-13), savestate branching (round 14), persistent internal goals (rounds 15-16), and further extensions (see regimes A-E in Table \ref{tab:regimes}).

\begin{figure*}[t]
\centering
\includegraphics[width=0.99\textwidth]{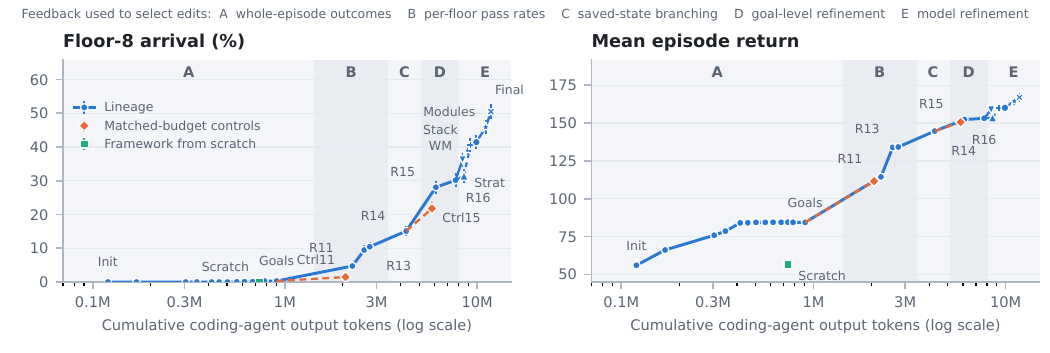}
\caption{Policy optimization on 2048 held-out test worlds. The 25 checkpoints share
ancestry and were selected before evaluation. The shading marks the feedback
regime. Error bars are (Wilson) 95\% intervals for arrival and normal intervals for mean return. Orange points are single controls with an equal-compute budget, and the green point starts from scratch.}
\label{fig:lineage}
\end{figure*}

To compare every stage fairly without selection bias, we froze all 25 checkpoints after optimization and evaluated them on a held-out set of 2048 fresh worlds (Appendix \ref{app:lineage}). Figure \ref{fig:lineage} plots final floor arrival and mean episode return against cumulative coding agent output tokens (11.8M tokens total, including reasoning and rejected branches).

\begin{table*}[t]
\centering\footnotesize
\begin{tabular}{@{}lrrrrl@{}}
\toprule
Frozen comparison & Worlds & Before & After & Gain & 95\% interval\\
\midrule
Goal-level: R14 to R16 & 2048 & 13.82 & 31.45 & +17.63 & [+15.26, +19.99]\\
World model & 2048 & 31.88 & 37.55 & +5.66 & [+2.97, +8.36]\\
Modules over composition & 4096 & 40.28 & 45.09 & +4.81 & [+3.08, +6.54]\\
Recursive refinement & 1024 & 44.43 & 48.24 & +3.81 & [+0.71, +6.91]\\
\bottomrule
\end{tabular}

\caption{Selected compositions on separate frozen cohorts. Before and after are
floor-8 arrival percentages, and gains and intervals are in percentage points.
Each row has its own incumbent, candidate and cohort.}
\label{tab:frozen}
\end{table*}

\paragraph{Main optimization trajectory.}
Under feedback regime A, mean return rose from 56 to 84 in four rounds and then plateaued there for six more rounds, leaving floor-8 arrival stuck at 0.29\%.
Switching to structured feedback broke this plateau at each stage: per-floor pass rates (B) lifted floor-8 arrival to 10.45\% by round 13, saved-state branching (C) reached 15.09\% in round 14, and goal-level refinement (D) jumped to 28.08\% in round 15 and 30.18\% in round 16. Extending into model and module refinement (E) pushed arrival to 36.23\% with the world model and shield, 45.90\% with cross-floor modules, and 50.49\%  after recursive module refinement.

\paragraph{Control experiments.}
Because the main optimization run is sequential, we ran three control experiments to test what happens when certain components of the cascade are removed:
\begin{enumerate}
    \item \emph{Can agents discover floor-level feedback on their own?} Restarting from the regime A plateau with round 11's agent setup and same token budget, but stating only the final-floor objective. The agent (Ctrl11) independently invented per-floor pass-rate evaluation within 75k tokens, but only achieved 1.46\% arrival (vs. 4.74\% for round 11 in the main run).
    \item \emph{Can agents discover cross-floor goal refinement on their own?} Restarting from round 14 with round 15's agents and same budget, but omitting our hierarchical goal framework. The Ctrl15 agent kept partitioning work by floors rather than cross-floor goals. It reached 21.78\% arrival, falling short of round 15's 28.08\%.
    \item \emph{Can we skip the early cascade and start directly at the goal level?} Given the game engine and the full goal-refinement harness from scratch with no initial policy (Scratch), an agent spent 737k tokens (matching round 10 budget) but achieved 0\% floor-8 arrival: 31 candidate edits went unpromoted because zero floor-8 arrivals occurred to score continuations. Advanced continuation feedback requires an incumbent that already reaches the endpoint occasionally, which the earlier regimes provide, thus motivating our cascade feedback design.
\end{enumerate}

In addition to the common cohort, Table \ref{tab:frozen} reports held-out frozen comparisons conducted at the end of each major stage on fresh world cohorts, confirming consistent positive gains at every step. Appendix~\ref{app:diagnostics} audits the candidate edits behind these checkpoints, and shows that local measures such as next-floor arrival are a poor guide to final-floor gains.

\subsection{Factorization and Execution}
\label{sec:parity}

Recall from Section \ref{sec:factorization} that the factorization is only valid if $I(p_0)$ and the specialized binary $t_0$ behave exactly as the policy they come from. To verify this, we did not take the Final policy (Fig. \ref{fig:lineage}) directly, but a descendant which was further refined for boss victory (Appendix \ref{app:boss}; wins went from 8.94\% to 28.37\% with the same rate of floor-8 arrivals) and later ported to C++ so as to make it cheaper to run, always under checks of action and skill parity.

\begin{figure}[t]
\centering
\definecolor{tealblue}{HTML}{367588}
\begin{tikzpicture}[font=\scriptsize,
  box/.style={draw=black!75, rounded corners=1pt, minimum height=4.6mm, minimum width=15mm, align=center, inner sep=1pt},
  arr/.style={-{Stealth[length=1.3mm]}, semithick, draw=black!75},
  lab/.style={font=\tiny, anchor=north, inner sep=0.5pt}]
\newcommand{\fzbar}[3]{%
  \fill[tealblue] ($(#1.south)+(-6mm,-2.4mm)$) rectangle ++(#2*12mm,1.3mm);
  \draw[black!75, line width=0.3pt] ($(#1.south)+(-6mm,-2.4mm)$) rectangle ++(12mm,1.3mm);
  \node[lab] at ($(#1.south)+(0,-2.8mm)$) {#3};}
\node[box] (c) {C policy};
\node[box, right=5mm of c] (sp) {C speed-up};
\node[box, right=5mm of sp, fill=tealblue!15] (cpp) {C++ port $t$};
\draw[arr] (c) -- (sp); \draw[arr] (sp) -- (cpp);
\fzbar{c}{0.20}{1$\times$}
\fzbar{sp}{0.79}{2 agents, 3.9$\times$}
\fzbar{cpp}{1}{4 agents, 4.9$\times$}
\node[font=\tiny, anchor=west] at ($(cpp.south east)+(1mm,-1.75mm)$) {speed vs.\ C policy};
\node[box, minimum width=6mm, fill=tealblue!15, below=7.6mm of c.south west, anchor=north west] (t) {$t$};
\node[box, right=5mm of t] (s1) {Stage 1};
\node[box, right=5mm of s1] (s2) {Stage 2};
\node[box, right=5mm of s2] (out) {$I(p_0)$, $t_0$};
\draw[arr] (t) -- (s1); \draw[arr] (s1) -- (s2); \draw[arr] (s2) -- (out);
\fzbar{s1}{0.51}{1 agent, 51\%}
\fzbar{s2}{0.95}{3 agents, 95\%}
\node[font=\tiny, anchor=west] at ($(s2.south east)+(1mm,-1.75mm)$) {floor refs.\ in $p_0$};
\end{tikzpicture}
\caption{From the C policy to the factorization. Top: the port, with the speed per decision relative to the original C policy. Bottom: the extraction from the C++ target $t$, with the share of literal floor references held by $p_0$ after each stage. Every step has to pass the parity check.}
\label{fig:factorization}
\end{figure}
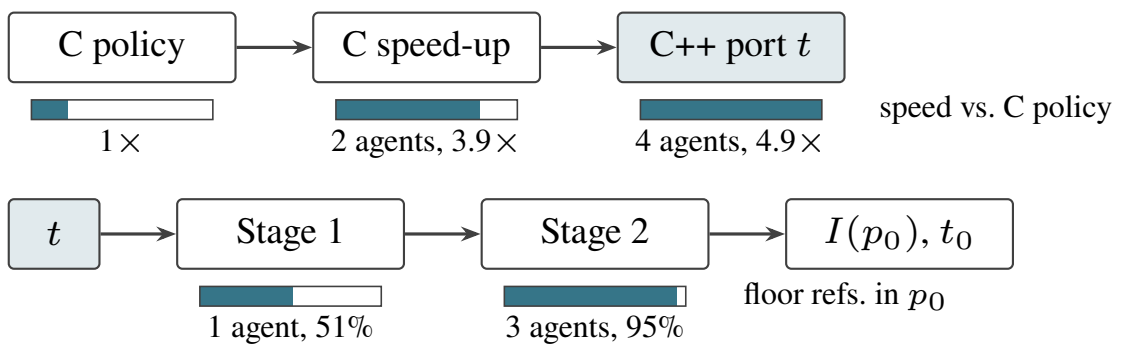

\paragraph{Parity check.}
The port and the extraction rely on the same check. We ran the reference policy once on a fixed set of worlds and recorded the action and the skill label that it chose at every decision. A candidate was then run on the same worlds from reset, and its action and skill were compared with the recorded ones decision by decision: an episode fails at the first mismatch, or if its length differs. Since the engine is deterministic given the world seed, two policies that agree at every decision go through exactly the same states. There are 64 development worlds (3.2M decisions), which the agents could run themselves, and 256 held-out worlds (12.4M decisions), which they never saw and were only used to confirm the merged result.

\paragraph{Port and extraction process.}
Both steps were carried out by coding agents (Claude Code with \texttt{claude-opus-5-5}), and every change had to pass the parity check (Figure~\ref{fig:factorization}). For the port, two parallel agents first made the C policy 3.9 times faster per decision; then, starting from a plain C++23 recompilation (which by itself gave no gain), three agents pushed time, memory and compile-time checks, respectively, and a fourth one combined the first two into the target $t$, which is 4.9 times faster than the original C policy. The extraction had two stages. In the first one, a single agent designed the rule language and its two evaluators, and moved the rule table of the policy to $p_0$, which at that point held 51\% of the literal floor references of the code (e.g. \texttt{level==5}). In the second one, three parallel agents moved the remaining strategy routines of supply, combat and planning, after which $p_0$ holds 95\% of the floor references and the ones left in the decision path of $I$ are game facts. Overall, the port took about four hours and 1.5M output tokens, and the extraction less than two hours and 0.6M (details in Appendix \ref{app:factorization}, prompts in Appendix \ref{app:prompts}).

After extraction, $p_0$ has 149 ordered rules and 152 helper definitions in seven files. We ran $I(p_0)$, $t_0$ and the C++ target on 64 development and 256 held-out worlds, a total of 15,564,565 decisions for each variant, and at every single one of them the action and the skill label coincided with those of the target (details in Appendix~\ref{app:factorization}). In terms of cost, a decision takes 9.5 $\mu$s with $t_0$ and 19.7 $\mu$s with $I(p_0)$, while the target needs 9.4 $\mu$s; so the overhead that interpretation adds is recovered by specialization.

\subsection{Goal Generalization}
\label{sec:generalization}
Our last experiment tests whether a compiler searched on short goal programs can be reused on new programs, including longer ones.

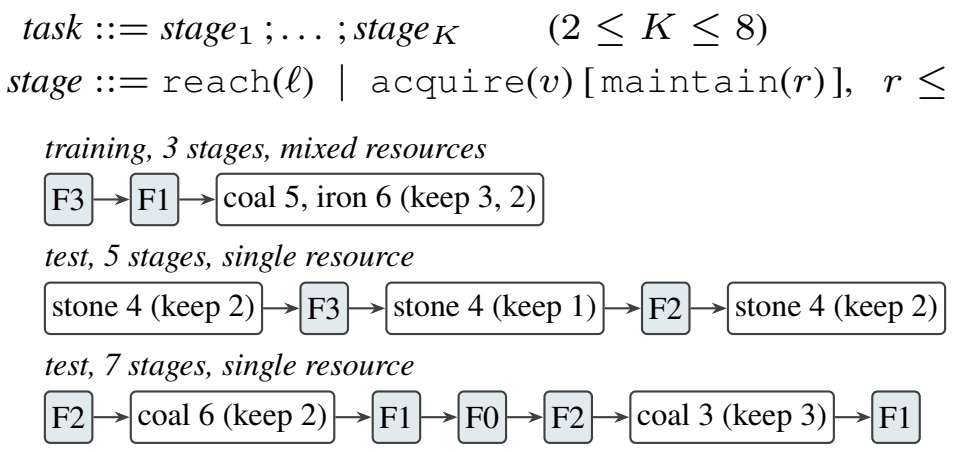
\begin{figure}[t]
\centering
\definecolor{tealblue}{HTML}{367588}
{\scriptsize
\begin{tabular}{@{}r@{\;}c@{\;}l@{}}
\textit{task} & $::=$ & \textit{stage}$_1$\,;\,\dots\,;\,\textit{stage}$_K$ \qquad ($2\le K\le 8$)\\
\textit{stage} & $::=$ & \texttt{reach}($\ell$) $\;\mid\;$ \texttt{acquire}($v$) [\,\texttt{maintain}($r$)\,], \; $r\le v$
\end{tabular}}

\vspace{1.5mm}
\begin{tikzpicture}[font=\tiny,
  rc/.style={draw=black!75, fill=tealblue!15, rounded corners=1pt, inner sep=1.3pt, minimum height=3.4mm},
  ac/.style={draw=black!75, rounded corners=1pt, inner sep=1.3pt, minimum height=3.4mm},
  arr/.style={-{Stealth[length=1mm]}, draw=black!75},
  lab/.style={anchor=south west, inner sep=0pt, font=\tiny\itshape}]
\node[lab] at (0,0.5mm) {training, 3 stages, mixed resources};
\node[rc, anchor=north west] (a1) at (0,0) {F3};
\node[rc, right=2.4mm of a1] (a2) {F1};
\node[ac, right=2.4mm of a2] (a3) {coal 5, iron 6 (keep 3, 2)};
\draw[arr] (a1) -- (a2); \draw[arr] (a2) -- (a3);
\node[lab] at (0,-6.7mm) {test, 5 stages, single resource};
\node[ac, anchor=north west] (b1) at (0,-7.2mm) {stone 4 (keep 2)};
\node[rc, right=2.4mm of b1] (b2) {F3};
\node[ac, right=2.4mm of b2] (b3) {stone 4 (keep 1)};
\node[rc, right=2.4mm of b3] (b4) {F2};
\node[ac, right=2.4mm of b4] (b5) {stone 4 (keep 2)};
\foreach \x/\y in {b1/b2, b2/b3, b3/b4, b4/b5} \draw[arr] (\x) -- (\y);
\node[lab] at (0,-13.9mm) {test, 7 stages, single resource};
\node[rc, anchor=north west] (c1) at (0,-14.4mm) {F2};
\node[ac, right=2.4mm of c1] (c2) {coal 6 (keep 2)};
\node[rc, right=2.4mm of c2] (c3) {F1};
\node[rc, right=2.4mm of c3] (c4) {F0};
\node[rc, right=2.4mm of c4] (c5) {F2};
\node[ac, right=2.4mm of c5] (c6) {coal 3 (keep 3)};
\node[rc, right=2.4mm of c6] (c7) {F1};
\foreach \x/\y in {c1/c2, c2/c3, c3/c4, c4/c5, c5/c6, c6/c7} \draw[arr] (\x) -- (\y);
\end{tikzpicture}
\caption{Task language and three of the generated programs. F$\ell$ is \texttt{reach}($\ell$); the other boxes are acquisitions, with the reserve to keep in parentheses.}
\label{fig:tasks}
\end{figure}

\paragraph{Tasks.}
The tasks were produced by a generator which was fixed before running any search: 16 training and 8 validation programs, of 2--4 stages each, and a test suite of 48 programs. The latter has 16 programs for each of the lengths 2--4, 5--6 and 7--8 and, within each length, six are navigation tasks, five are single-resource and five mixed-resource (Figure~\ref{fig:tasks} shows the task language and three of the programs). Floors range over 0--3 and resources over wood, stone, coal and iron, with programs differing in the order, composition and stock requirements of their stages; none of them had been used during policy development. We only checked syntax and cumulative bounds, i.e. no task was discarded for being unreachable or too hard.

\paragraph{Shared compiler search.}
For the shared compiler we launched five independent searches, each one a fresh \texttt{gpt-6.1-sol} context that started from the same starter emitter, diagnostics and allowed training feedback (the prompt is in Appendix~\ref{app:prompts}). Each search evaluated exactly three candidate emitters on the 16 training tasks, using 4 worlds per task (192 episodes). The parent run then scored the three candidates on the 8 validation tasks and 8 separate worlds (192 episodes more) to pick one, and this validation feedback was never returned to the coding agent. We evaluated the five selected compilers, not just the best one. No search had access to the test specifications, to the test outcomes, or to the other four searches.

\paragraph{Baselines.}
To separate the effect of the emitted controller from that of the constraints, we used three fixed controls. The first one, \texttt{Original}, is simply $p_0$: it ignores the task, which we score independently. \texttt{Guards} is $p_0$ with the resource-maintenance constraints, but without goal controller or navigation barrier. \texttt{Reference} adds to \texttt{Guards} a prescribed reactive sequence controller (it travels to the active floor, or obtains the tools and gathers the stock). Our method, \texttt{Shared}, uses the emitters selected above. Finally, \texttt{Per-task} tells us what is gained by knowing the task in advance: given the specification of a new task, it runs a search just for it in the same emitter interface, with a fresh coding-agent context, three candidates on 8 tuning worlds and selection on 8 other worlds. We ran \texttt{Per-task} on a fixed subset of 12 tasks (for each length stratum, one navigation, one single-resource and two mixed-resource tasks). So \texttt{Per-task} can specialize to its specification and \texttt{Shared} cannot. Progress semantics and hard constraints are the same in all guarded methods.

\paragraph{Evaluation.}
Each method was run on the same 32 fresh test worlds per task (so comparisons are paired across methods), with a horizon of 50,000 actions. In total there are 12,672 episodes: 1,536 per fixed control, 7,680 for the five shared searches and 384 for \texttt{Per-task}. Training, selection and test cohorts are disjoint, and both the sources and the analysis were frozen before we saw any feedback from the test worlds. Our primary contrast is the mean success of \texttt{Shared} minus that of \texttt{Guards}. Its interval comes from a crossed bootstrap over coding searches, worlds and tasks (Appendix~\ref{app:statistics}). The other intervals are descriptive and are not adjusted for multiple comparisons.

\begin{figure}[t]
\centering
\includegraphics[width=\columnwidth]{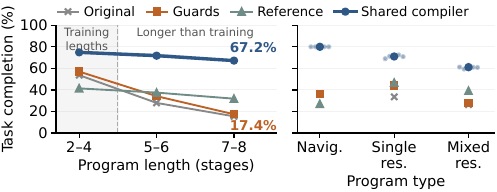}
\caption{Transfer to unseen goal programs, on 32 common worlds per task. The
shading marks the lengths used for training and validation. Blue circles average
five searches, and faint lines and dots are the individual searches (not
confidence intervals). Since the horizon is fixed, longer programs also demand
more actions.}
\label{fig:transfer}
\end{figure}

\paragraph{Overall completion.}
Table~\ref{tab:compiler} has the results on the 48 test tasks. The shared compilers complete \FCSharedSuccess\% of them on average, against \FCGuardsSuccess\% for \texttt{Guards} and 36.91\% for \texttt{Reference}. The primary difference is +35.10\% percentage points, and \texttt{Shared} needs 13,727 fewer failure-penalized steps than \texttt{Guards}. No guarded method has a maintenance violation, while \texttt{Original} has 164. The guards are thus enough to enforce the constraints, but they explain only a small part of the completed tasks.

\begin{table}[t]
\centering\footnotesize
\begin{tabular}{@{}lrrr@{}}
\toprule
Method & Success & Violations & Steps$^\dagger$\\
\midrule
Original & 32.36\% & 164 & 38,358\\
Guards & 36.20\% & 0 & 37,258\\
Reference & 36.91\% & 0 & 35,104\\
Shared compiler & 71.30\% & 0 & 23,531\\
\bottomrule
\end{tabular}

\caption{Results on all 48 goal-program test tasks. \texttt{Shared} averages five searches,
and its violation counts pool their trials. Each fixed control has 1,536 trials
and \texttt{Shared} has 7,680. $\dagger$Failures are charged the full 50,000-action
horizon.}
\label{tab:compiler}
\end{table}

\paragraph{Longer programs and task types.}
Figure~\ref{fig:transfer} splits the results by program length. \texttt{Shared} succeeds on 74.88\%, 71.80\% and 67.23\% of the short, medium and long programs, while \texttt{Guards} drops from 57.03\% to 34.18\% and 17.38\%. Recall that the search only saw programs with 2--4 stages, so the compiler transfers to programs longer than the ones used to optimize it. This does not depend on one particular coding run either: over the five searches, overall success goes from 70.70\% to 72.07\% (sample standard deviation of 0.59 points). Tasks with interacting resources are still the hardest ones. \texttt{Shared} completes 80.03\% of the navigation, 71.04\% of the single-resource and 61.08\% of the mixed-resource tasks (\texttt{Guards}: 36.46\%, 44.17\% and 27.92\%). During development we saw that the reserves often blocked crafting, and the emitted controllers adjust the buffers and the routing for this.

\paragraph{Comparison with per-task search.}
For the 12-task subset we obtain 69.48\% with \texttt{Shared} and 65.62\% with \texttt{Per-task}. This subset is heavy in resource tasks. Hence a frozen compiler, deployed without any tuning for the task, is on par with one round of model-guided search per specification.

\paragraph{Execution checks.}
We conclude with some sanity checks. We ran 24 tests on semantics, resource costs, generation, auditing and analysis. Interpreted and specialized execution agree on 28 cases (262,588 decisions). On navigation tasks without reserves, \texttt{Original} and \texttt{Guards} give the same 576 paired episodes up to the instrumentation of progress, and the native monitor does not depend on the controller's memory.

\subsection{Costs and Limitations}
\label{sec:limits}

A reusable compiler changes the cost of every future task. After selection, \texttt{Shared} does not need rollout tuning or model calls for a new task, although emission and native compilation still take some time. Conceptually, the costs can be written as $B_{\rm shared}(N)=B_{\rm develop}+\sum_{i=1}^N b_{\rm compile}(g_i)$ and $B_{\rm direct}(N)=\sum_{i=1}^N b_{\rm search}(g_i)$, where both are conditional on the common machine and base policy (whose development and factorization also have to be paid for by the whole workflow). In our experiments, each shared search uses 640 logical feedback episodes, or 7.76--8.36 million steps, and this counts the shared baselines in full for every search. The twelve per-task searches use 960 episodes in total, which is 12.22 million steps. In output tokens, a shared search used 13.4k on average and a per-task search 7.5k. Each new task thus costs \texttt{Per-task} 80 episodes (about 1.02 million steps) and 7.5k output tokens, and costs \texttt{Shared} nothing. Since \texttt{Shared} is on par with \texttt{Per-task} on the common subset, a shared search pays for itself after eight tasks in feedback episodes and after two in output tokens, and the test suite alone has 48. Appendix~\ref{app:statistics} has the full accounting.

\paragraph{Limitations.}
First, the lineage is a single adaptive development history. The authors introduced each regime when the diagnosis suggested it, so the feedback is confounded with ancestry, though several controls experiments were performed. Second, the synthesized interpreter is a Craftax-specific implementation with a lot of inherited knowledge, and parity is an empirical check on held-out worlds. Third, goal generalization is restricted to a supplied language and one game engine. The composition and progress semantics come from the infrastructure and each per-task baseline is a single search.

\section{Related Work}

Programmatic reinforcement learning searches over structured policy spaces \citep{pirl,leaps,hprl}, and language-model-guided search adds code proposals made by a language model \citep{llmgs}. Code as Policies generates executable control from instructions, Voyager develops reusable skills, and WorldCoder writes executable world models \citep{codepolicies,voyager,worldcoder}. In our case, the object that is deployed is a frozen program-to-program map over a supplied goal language, and the language model only takes part in development.

FunSearch and AlphaEvolve combine edits written by a model with automated evaluation \citep{funsearch,alphaevolve}. Dense feedback can expose failures that look the same under a scalar score \citep{gallego}. Our feedback cascade also changes the starting state of the comparison, the scope of the code and the continuation. Saved-state branching is related to restoring states for exploration \citep{goexplore}, although here the paired continuations are used to evaluate local program edits against the chosen endpoint.

Options and hierarchies of machines formalize temporally extended behavior and partial control programs \citep{options,hams}. MAXQ distinguishes the recursive optimality of a subtask from the performance of the whole hierarchy, and ALisp takes into account what follows a subroutine through its exit value \citep{maxq,alisp}. Our continuation formulation applies the same distinction to the edits of coding agents and to handoffs between programs. Policy sketches and reward machines also expose task structure for composition \citep{sketches,rewardmachines}. In our case the task semantics and the monitor are given, and the code search adapts how the controller uses the skills it already has. The resource guards are similar to runtime shields \citep{shielding}, but without a synthesized guarantee on the winning region.

Goal-conditioned reinforcement learning trains a single network over a goal space, and the generalization comes from universal value function approximators and hindsight relabeling \citep{uvfa,her,gcrl}. In Craftax, \citet{matthews2026goal} train neural policies over a discrete set of semantic goals by outputting values for all goals at once (LEO). Our goal compiler can be seen as a programmatic family of goal-conditioned policies: the goal enters through code generation over a fixed machine instead of as a network input or output head, the generated program can be inspected and runs without a model, and generalization is tested on multi-stage compositions with constraints that are longer than any seen in training.

\section{Conclusions}
In order to be useful, a program policy first has to acquire competence and then expose it in a form that new objectives can reuse. In this work, feedback that becomes progressively more structured took the policy from a final floor that was almost never reached to 50.49\% arrival. The continuation view explains why gains in local completion and in local prediction have to be separated from their downstream value. Factoring the resulting policy into a machine and a rule program opens a second search space, the one of goal compilers. The five frozen compilers that we obtained transfer from short training programs to longer unseen compositions, without model calls for the new tasks, and a single compiler search pays for itself after eight tasks when compared with searching for each task separately. Overall, our contribution is a connected neuro-symbolic workflow for developing and reusing executable control, in which the model searches over programs and the program acts.

\paragraph{AI-Use Disclosure.}
Coding agents are part of the method studied in this paper. Claude Code
(\texttt{claude-opus-5-5}), Codex (\texttt{gpt-6-astra}) wrote and evolved the policies, the factorization into $I$ and $p_0$, and the goal-compiler emitters. The authors set the harness, (based on code from previous research), objectives, designed the feedback regimes and compiler setups (Appendix \ref{app:protocols}). AI assistants were used to write analysis and plotting scripts, and to draft and edit parts of the manuscript. The authors take full responsibility for the content of the paper.

\label{mainend}
\clearpage
\setlength{\parskip}{\savedparskip}
\bibliography{feedback_goal_compilers}
\clearpage
\appendix

\section{Observation Space}
\label{app:observation}

\begin{figure}[!b]
\centering
\includegraphics[width=\columnwidth]{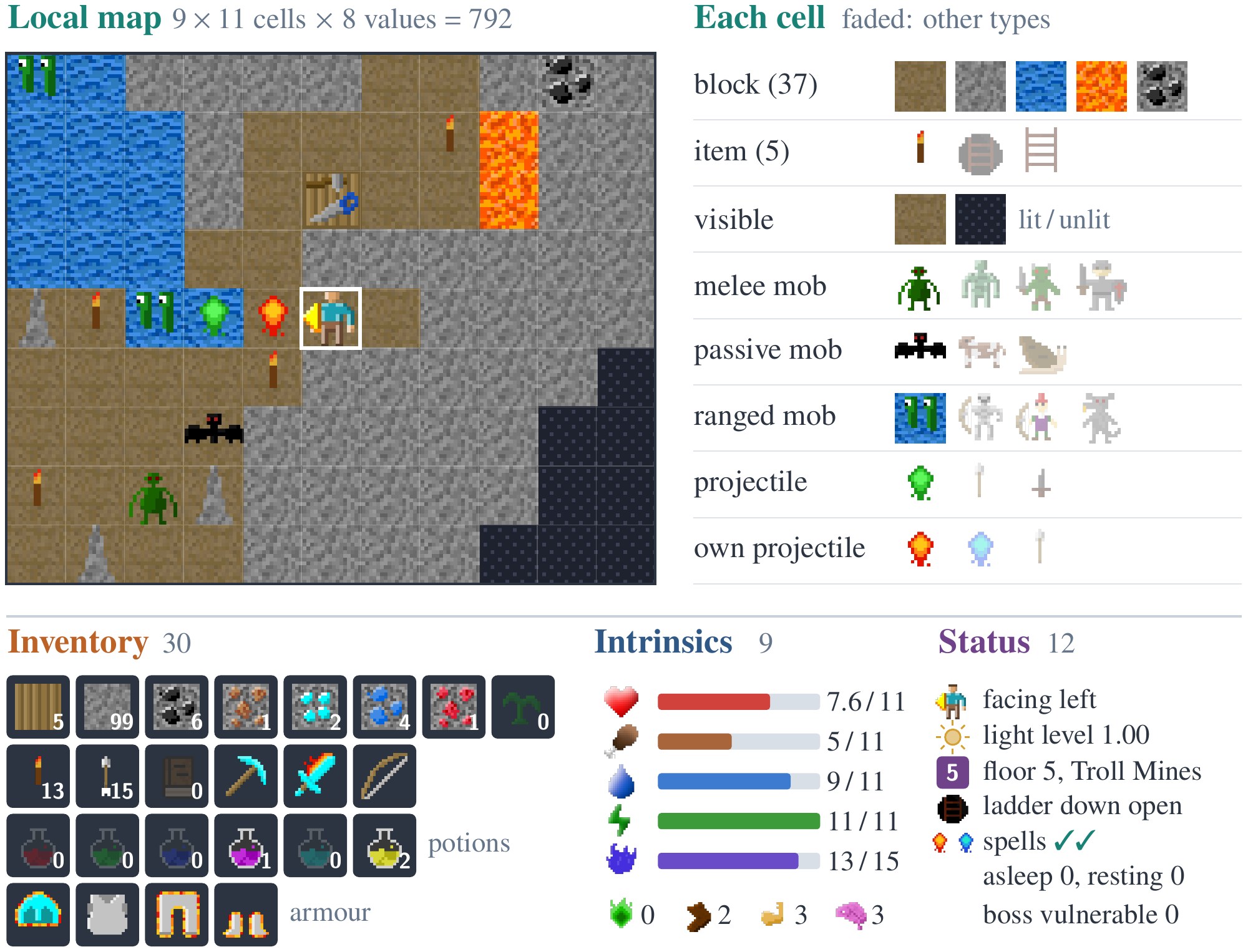}
\caption{A single observation, at step 16,855 of an episode played by the developed
program policy in the Troll Mines (floor 5). Top left: the $9\times11$ local
map, rendered only from the observation with the Craftax sprites
\citep{craftax}. The outlined cell is the agent, and the dark dotted cells are
unlit. Top right: the eight values of each cell, with the sprites that appear in
this view and, faded, the other types of the same class. Bottom: the 51 scalars
at this step, where the bars show each intrinsic with respect to its maximum.}
\label{fig:observation}
\end{figure}

Figure~\ref{fig:observation} shows all the information that a program receives at one step.
All five creature classes are present in this view: a troll, a bat, two deep
things, a slimeball in flight and a fireball thrown by the agent itself.

In total, the observation has 843 values. Most of them come from the local map,
which covers the $9\times11$ cells around the agent. For each cell there are
eight values. The first one is the block type (there are 37 types), the second
one is the item that lies on the cell, if any (a torch or a ladder), and the
third one is a visibility bit. The other five are slots, one for each creature
class, that hold the type of the creature that is in the cell. The five classes
are melee, passive and ranged creatures, the projectiles thrown by creatures,
and the projectiles thrown by the agent. When a cell is unlit its visibility bit
is cleared, and in that case it does not carry any block, item or creature.

The remaining 51 values are scalars, and they describe the inventory, the
intrinsics and the status of the agent. The inventory takes 30 of them. There
are eight materials, then the torches, arrows and books, then the levels of the
pickaxe, the sword and the bow together with the enchantments of the sword and
of the bow, then six types of potion, and finally the four armour pieces with
their enchantments. For the intrinsics there are nine values (health, food,
drink, energy, mana, experience, dexterity, strength and intelligence). The
status uses the last twelve values: four indicators for the facing direction,
the light level, one flag for sleeping and one for resting, the two spells that
can be learned, the current floor, and two more flags that say if the descending
ladder of the floor is open and if the boss is vulnerable.

The program never reads the hidden state of the engine. So whatever it knows
about earlier floors, ladder locations or resource sites has to come from the
memory that it builds out of these observations, as in Eq.~\eqref{eq:memory}.

\section{Experimental Details}
\label{app:details}

\subsection{Cohorts, Selection and Researcher Roles}
\label{app:protocols}

\paragraph{Policy development.}
An episode counts as an arrival when the deepest floor it reaches is at least 8.
The native return adds up the achievement and health terms of the evaluator. For
the mean achievement coverage we take the arithmetic mean of the fraction of the
67 distinct achievements obtained (note that this is not the weighted aggregate
score of the benchmark). The evaluations last 100,000 actions and they continue
after arrival, so a later boss victory can be recorded too. In this way,
arrivals, victories and native return come from the same complete episode, but
each of them describes a different outcome of it.

For the common-cohort evaluation, the command-line seed is 81,000,000, and we use
2,048 episode indices with the archived SplitMix64 seed scheme. This evaluation
was only used to place checkpoints chosen beforehand, and its source list was
fixed before any outcome was seen. The plan was originally 4,096 worlds. It was
reduced to 2,048 for runtime reasons, before any completed chunk had exposed
outcomes, and the dated amendment is kept. This cohort is not the same as the
development banks, nor as the frozen campaign tests of the selected programs.

In rounds 15--16 the development comparison used 512 worlds and seed
53,000,000. The entry banks for round 15 were captured from round 14, on 16,384
worlds with seed 52,000,000, while later screening and confirmation used other
development streams. The frozen comparison of round 14 against round 16 used
seed 57,000,000 on 2,048 worlds. It was planned with 8,192 worlds, but again we
had to reduce it for runtime, before looking at the final outcomes. The partial
outputs were excluded, and both arms were started again on the fixed cohort. The
test finds 517 worlds gained and 156 lost for final-floor arrival. We keep all of
these streams and amendments as separate records.

One thing to keep in mind is that a continuation is not an independent world,
because several entries and random replicas can come from the same source world.
When regions are merged, we check the candidate of the agent that owns a region
against the exact integrated source up to the boundary of that region. In the
round-15 merge, all 144 pairs that we sampled have the same action hash, length,
deepest floor and death. These checks tell us that the code composes correctly
under the ownership convention.

\paragraph{Agents and interventions.}
Up to round 16, the session logs have 9,916 model responses and 7.73 million
output tokens, where reasoning is included and each response is counted once.
After that, the kit-goal, world-model and module runs add about 0.82, 0.67 and
1.88 million tokens, and the recursive follow-ups add 0.71 million more, which is
how we get to the 11.80 million tokens of the arrival-focused axis. Rejected
branches are included in these totals, at the scope where they were recorded.
The later work on the boss, on execution optimization and on the compiler search
is not included.

As for the agent systems, we alternated between Claude Code running \texttt{claude-opus-5-5}, and Codex 
\texttt{gpt-6-astra} agents. The authors were the ones who set the objectives,
introduced the ideas of branching, hierarchical goals and models/modules, allowed
delegation, changed how often integration happened, and decided when to stop.
The agents wrote and evaluated the candidate code. So what this study measures is
a research process with coding agents that is directed by humans. The
matched-budget controls keep the specified checkpoint, agent system and budget,
and each of them is just one search.

\paragraph{Goal-program study.}
Table~\ref{tab:goalcohorts} lists the world cohorts, which are all distinct. The
generator fixes the tasks, their syntax and bounds, uniqueness up to names, and
the quotas for length and resources, and all of this happens before any feedback.
In the test set, each length stratum has six navigation programs, five
single-resource programs and five mixed-resource programs, and the mixed ones use
the pairs wood/iron, wood/stone and coal/iron. For per-task synthesis we use a
fixed subset, with one navigation, one single-resource and two mixed-resource
programs from each stratum. In each stratum, three of these tasks have the lower
length and one has the upper length (2/3, 5/6 and 7/8), so the subset does not
have any task of length 4.

\begin{table*}[t]
\centering\small
\begin{tabular}{@{}lrrrr@{}}
\toprule
Purpose & Task programs & Stages & Worlds/task & Seed base\\
\midrule
Shared training & 16 & 2--4 & 4 & 2,000,000,000\\
Shared validation & 8 & 2--4 & 8 & 2,010,000,000\\
Held-out test & 48 & 2--4 / 5--6 / 7--8 & 32 & 2,020,000,000\\
Per-task tuning & 12 fixed specifications & 2--8 & 8 & 2,030,000,000\\
Per-task selection & Same 12 & 2--8 & 8 & 2,040,000,000\\
\bottomrule
\end{tabular}
\caption{Cohorts of the goal-program study. Every episode has a horizon of
50,000 actions. Shared validation uses new tasks as well as new worlds, while
per-task selection keeps the known specification and only changes the worlds.
The shared searches cannot inspect the specifications of the validation and test
tasks during the search.}
\label{tab:goalcohorts}
\end{table*}

Each search archives its three candidates, together with the training results,
the closure hashes and the selection records. Repairs that only fix syntax are
allowed before the evaluated source is frozen, but once code has been evaluated
it cannot change anymore. In validation, the choice is made by number of
successes, then by failure-penalized steps, and then by taking the earlier
candidate. Writes for selection are serialized, and the search agents never get
the validation feedback back. The final selected source is an exact copy of one
of the archived candidates. The five contexts are separate, but they share the
starting sources and the diagnostics.

All shared searches start from the same starter emitter and see the same
baseline feedback. The complete archive also keeps the rows from the evaluation
of the starter. We count these rows in the feedback costs, but they are not part
of the 12,672 held-out episodes analyzed here, which cover the three fixed
controls, the five selected shared compilers and \texttt{Per-task}.

\subsection{Statistical Analysis and Costs}
\label{app:statistics}

\paragraph{Policy and continuation comparisons.}
For the marginal rates of arrival and victory, the intervals are Wilson 95\%
intervals. When we compare two policies from reset, each world $w$ gives a paired
difference $d_w\in\{-1,0,1\}$, the candidate minus the incumbent, and the
interval for the mean difference is $\bar d\pm1.96s_d/\sqrt{W}$. We use exact McNemar tests on the
counts of discordant successes. For example, on the common cohort the difference
between Final and Modules has 301 worlds gained and 207 lost, and the separate
round-16 comparison has 517 gained and 156 lost.

Things are a bit different when a source world has several continuations. Let
$d_{wi}$ be a paired difference in success, $n_w$ the number of pairs that come
from world $w$, $N=\sum_w n_w$ and $S_w=\sum_i d_{wi}$. Then the pair-weighted
estimate and its clustered standard error are
\begin{equation}
 \bar d=\frac{\sum_w S_w}{N},\quad
 \widehat{\mathrm{SE}}(\bar d)=
 \sqrt{\frac{W}{W-1}\frac{\sum_w(S_w-n_w\bar d)^2}{N^2}}.
\end{equation}
Note that with unequal cluster sizes this gives the same weight to every
continuation, rather than to every world. The two runs in a pair start from the
same captured state and initial random stream, but later random draws can
diverge once their actions differ. We treat the adaptive development tests and
the secondary analyses as nominal, and we do not adjust them for multiple
comparisons.

\paragraph{Compiler-search analysis.}
For the primary estimand, we average the success over three things: the five
coding searches, the task mixture that we declared in advance, and the 32 paired
worlds. The interval is obtained with a bootstrap of 10,000 draws. In each draw
we resample the five searches, and we also resample one list of 32 world
indices, which is then used for every task and every method. Within each cell of
length and resource kind we resample the tasks, but the cells keep their weights
(six navigation, five single-resource and five mixed-resource tasks per length
stratum). We use 20261024 as the bootstrap seed and report percentile 95\%
intervals. When an interval is given for one search alone, the searches are not
resampled. We do not adjust the secondary contrasts, which are only meant to be
descriptive.

For the direct subset, the tasks are resampled inside the declared cells. Note,
however, that every navigation and single-resource cell of this subset has a
single task. The comparison is conditional on the one direct search we selected
for each task, which means that it leaves out the variability of direct search
within a task. We did not declare an equivalence margin.

To measure speed we use failure-penalized steps: the completion time if the task
succeeds, or 50,000 if it fails. We only use the raw environment steps to count
how much evaluation work was done, and the two behave quite differently. An agent
that dies early, for example, consumes few environment steps but still receives
the full failure penalty. This is why plots and comparisons of synthesis cost
should not mix the two measures.

\paragraph{Feedback budgets.}
Next, we count how much feedback each method used. A shared search sees
$3\times16\times4=192$ candidate episodes and $3\times8\times8=192$ selection
episodes, and it is also shown 256 baseline episodes, which makes 640 in total.
The baseline is shared by all the searches, so we only had to run it once. A
direct search sees $3\times8$ tuning episodes, $3\times8$ selection episodes and
32 baseline episodes. Putting the five shared searches and the twelve direct
searches together, 3,136 feedback episodes were physically run. Of these, 1,920
come from the shared candidates and their selection, 256 from the common baseline
and 960 from \texttt{Per-task}. We keep diagnostics apart from the feedback used for
adaptation: the eighteen common diagnostic and replay episodes of development
take 623,034 steps, and the parity validation takes 56 episodes and 525,176
steps.

Regarding time, the agent records give the observed elapsed time, the successful
compilation preflight and the wall time of each evaluated round. These numbers
include waiting, so they should not be read as uncontended model or CPU time.
Also, when a syntax repair failed, we did not time that attempt on its own.

As for tokens, the session logs give 67,229 output tokens for the five shared searches (reasoning included; between 11.3k and 16.5k each) and 22.9M input tokens, of which 22.4M were cached. The twelve per-task searches used 89,865 output tokens (5.8k to 11.3k each) and 18.0M input tokens, 17.2M of them cached. These counts do not include the orchestrating session.

\paragraph{Source integrity.}
Before any new training feedback, the compiler protocol freezes 94 input files,
and before the held-out test it freezes 37 more inputs with the selected sources
and the analysis. Evaluating the frozen compilers needs no model calls.

\section{Policy Structure, Factorization and Compilation}
\label{app:factorization}

\paragraph{Memory and boundaries.}
The program keeps its own maps, both local and global, together with the cells it
has visited, the ladders it knows about, the resource sites and what it estimates
about adversaries. Each internal goal has a frame that holds its identity and its
local state, but the persistent world memory is shared between parent and child.
We use a post-observation boundary convention, which means that the runtime
advances at most once for each new observation. Note also that completing a
resource trip is not the same as physically returning: a return event gives
control back to the caller even if the child has abandoned its resource
objective. If the agent dies, all the active invocations are unwound. Finally,
when a runtime is inserted it must preserve the actions that already existed,
and only after that can new behaviors be enabled.

The following schematic shows the timing convention. It is only an illustration,
and it does not replace the retained implementation:
\begin{lstlisting}[language=C]
observe_public_state();
update_world_and_resource_memory();
if (dead()) return_all(DIED);
else if (active_goal_finished_or_abandoned())
    return_to_parent(recorded_outcome);
/* Survival rules retain higher priority. */
execute_first_rule_that_emits_an_action();
\end{lstlisting}

\paragraph{The source bridge.}
Several steps separate the policy of the main lineage from the program that is
factored. The arrival-focused lineage ends at Final. We then refined Final for
the boss fight (Appendix~\ref{app:boss}) and froze the boss-refined policy as the
C baseline of the execution-cost study. That study rewrote it in C++ for speed
while keeping every autonomous action and skill label unchanged (the parity check
checks both), and we factor the C++ version. The extraction into the
rule language is the step that exposes the strategy as $p_0$ over $I$. Then the
composition study freezes 35 dependencies between the machine and the program,
and it checks the image of the base program.

After extraction, the program has 149 ordered rules, 152 helper definitions and
24 query answers, which are spread over seven files. The language supports typed
integer and float operations, local arrays, out-parameters and static call
frames. When it is deployed, the program calls the primitives of the machine
through exported interfaces, and the machine calls back to the program through
explicit queries whenever it needs a strategy parameter. There is also a
commanded-goal mode, which still has its own code, but we disable it in the
composition study.

\paragraph{Port and extraction process.}
We give here more details on the two steps of Figure~\ref{fig:factorization}. In the first part of the port, the two agents owned disjoint parts of the C code (the path search, and everything else), and a root agent merged their snapshots at checkpoints. The time per decision went from 47.6 to 12.4 $\mu$s. Recompiling this code as C++23 ran at exactly the same speed, so in the second part three agents started from the plain port and each one pushed a different property: time per decision (1.28 times faster), memory per instance (from 1.33 MB to 349 KB) and compile-time guarantees, where the compiler itself replayed the first 3,400 steps of a development episode and checked every action. A fourth agent combined the first two, and the result is the target $t$, with 350 KB per instance. Along the way the agents reported four bugs, none of which changes autonomous play.

As for the extraction, in the second stage the root agent first extended the interface in both directions: $p_0$ exposes four entry points (initialization, observation, action filter and commit), and $I$ may call $p_0$ back through queries whenever a generic routine needs a strategic parameter. Each of the three subagents owned one area of the code and had to keep parity on the development worlds at every snapshot, while the root agent merged the snapshots and reran parity, on the held-out worlds as well. The C++ policy has 510 literal floor references. After the first stage 266 of them are in $p_0$ and 259 in $I$; at the end, $p_0$ has 467 and the decision path of $I$ has 24, e.g. which floors are lit or where the boss is. On our side, we only proposed the two questions of the port (the inference cost, and what C++ adds), chose the policy to factor and launched the second stage.

\paragraph{Parity and specialization.}
The dynamic evaluator loads a typed image of the nodes. The specialized evaluator
does something different: it turns the node indices into template parameters and
the node data into constants. Since both of them use the same helper for each
operation, the arithmetic and the order of side effects are the same in the two.
Both builds of the final factorization
(interpreted and specialized) agree with the C++ target on 64 development worlds,
with 3,209,631 decisions, and on 256 held-out worlds, with 12,354,934 decisions.
To catch cases where the program depends by accident on old contents of the
scratch memory, we poison that memory first. The criterion that we use is strict:
the action and the skill have to be the same at every tested step, so having the
same aggregate reward is not enough.

The report also includes a timing trace, where we kept the best of three runs in
that session. The C++ target needs 9,438 ns per decision, interpretation needs
19,714 and specialization 9,493, which means that specialization gets back the
overhead added by interpretation. For the task-conditioned policies we did 28
more comparisons between interpreted and native execution. These match in rows
and action hashes over 262,588 decisions, on the training and tuning worlds that
were allowed once the source was closed.

\subsection{A Worked Goal-to-Program Example}
\label{app:example}

For this example we take a test program with four stages, and the emitter that
was selected in shared search 5. The reason for choosing it is that it has a
return, an acquisition and also a maintenance bound that applies later on. The
displayed source is the program that this emitter generates for the task.

\begin{figure}[h]
\centering
\begin{minipage}{\columnwidth}\footnotesize
\input{feedback_goal_compilers_assets/example.tex}
\end{minipage}
\caption{Translation of a concrete task, and the source that the selected
emitter produces for it. The coal columns show only part of the full tables,
which cover five resources. Milestone semantics belong to the fixed translation,
and the controller strategy belongs to the emitter.}
\end{figure}

The parser accepts floors from 0 to 8 and bounds from 0 to 20 for wood, stone,
coal, iron and diamond, but the generator in our experiment only uses floors 0--3
and never asks for diamonds. A reach stage is completed when the agent stands on
the requested floor, and an acquisition stage when all the named inventory bounds
hold. The monitor accepts at most one milestone transition per observation, and
only while the agent is alive. Items already in the inventory can count towards a
later acquisition. Maintenance starts once that acquisition is completed and lasts
until the task ends, using the coordinatewise maximum of the bounds completed so
far. The task fails if there is a violation, if the agent dies, or if the horizon
runs out first.

Going back to the example, the floor table is $[1,0,-1,2]$ ($-1$ is the acquisition stage), the coal stock column is $[0,0,6,0]$ and the hold column is $[0,0,2,0]$. Once the third milestone is reached, the bound of two coal remains active while the policy goes on to floor 2. Since the goal goes further than floor 1, the emitter in the figure sets \texttt{SR\_SIMPLE=0}, and in its branch for difficult goals more of the equipment preparation of the base program is kept. For the returns, it emits a clause for upward travel that reuses the navigation already in the machine. Other selected emitters have their own rules for reserves and routing.

\paragraph{Audit and monitor boundaries.}
What the emitter is allowed to see is the structure of the goal and the public
interfaces. The search rules forbid a few things explicitly: behavior that
depends on the name of the goal, access to world seeds or to results, external
I/O, replacing the common semantics, and writing into the task-progress fields.
We enforce these rules with static checks on the Python and TR code, and we also
reviewed the source manually. The monitor reads the native engine state on its
own and never trusts the progress that the policy reports, nor the completion
predicates that were emitted. The hard filter blocks resource spending that would
cross an active positive reserve, but it adds no navigation barrier, and all
guarded methods use this same filter.

\section{Additional Experimental Results}
\label{app:results}

\subsection{Complete Lineage}
\label{app:lineage}

Table~\ref{tab:alllineage} has the 25 retained checkpoints from the common-cohort
figure of the main text. The rows are grouped by role, and not all of them are
sequential descendants (for example, kit goals and the world model branch from
round 16 and are composed afterwards). The token coordinate of the final point
counts all five recursive follow-up campaigns, rejected branches included. The
two single-run controls and the run from scratch are in the main figure and in
the text, but they are not checkpoints of the lineage.

\begin{table*}[t]
\centering\small
\begin{tabular}{@{}llrrrr@{}}
\toprule
Checkpoint & Regime & Tokens (M) & Arrival (\%) & Wins (\%) & Return\\
\midrule
Initial & A & 0.12 & 0.00 & 0.00 & 56.0\\
R1 & A & 0.17 & 0.00 & 0.00 & 66.2\\
R2 & A & 0.30 & 0.00 & 0.00 & 75.9\\
R3 & A & 0.35 & 0.00 & 0.00 & 78.6\\
R4 & A & 0.42 & 0.00 & 0.00 & 84.0\\
R5 & A & 0.46 & 0.00 & 0.00 & 84.1\\
R6 & A & 0.50 & 0.00 & 0.00 & 84.4\\
R7 & A & 0.56 & 0.05 & 0.00 & 84.4\\
R8 & A & 0.62 & 0.10 & 0.00 & 84.5\\
R9 & A & 0.68 & 0.10 & 0.00 & 84.5\\
R10 & A & 0.74 & 0.29 & 0.05 & 84.5\\
Base & A & 0.78 & 0.29 & 0.00 & 84.4\\
Goal interface & A & 0.90 & 0.29 & 0.00 & 84.4\\
R11 & B & 2.24 & 4.74 & 0.20 & 114.5\\
R12 & B & 2.58 & 9.47 & 0.88 & 133.9\\
R13 & B & 2.76 & 10.45 & 0.88 & 134.2\\
R14 & C & 4.28 & 15.09 & 1.81 & 144.7\\
R15 & D & 6.10 & 28.08 & 5.08 & 152.3\\
R16 & D & 7.73 & 30.18 & 5.71 & 153.2\\
Kit goals & D & 8.55 & 31.45 & 6.84 & 153.5\\
World model & E & 8.40 & 36.23 & 3.27 & 158.8\\
Composition & D & 9.22 & 40.58 & 5.71 & 160.0\\
Regions & D & 9.91 & 41.46 & 4.64 & 160.1\\
Modules & E & 11.10 & 45.90 & 5.96 & 164.6\\
Final & E & 11.80 & 50.49 & 8.94 & 167.0\\
\bottomrule
\end{tabular}

\caption{All the arrival-focused checkpoints that were retained, evaluated on the
same 2,048 worlds. Tokens are the cumulative output of the coding agents, in
millions and including reasoning. Return is the mean native return given by the
evaluator. The regime column says how the feedback was organized in the stage
that produced the checkpoint. Ancestry and budget are not controlled from one row
to the next.}
\label{tab:alllineage}
\end{table*}

\subsection{Feedback Diagnostics}
\label{app:diagnostics}
This section gives more details about the candidate edits behind the checkpoints of Section~\ref{sec:development}. In particular, it shows why local measures of progress were not used to select them.

\begin{figure}[t]
\centering
\includegraphics[width=\columnwidth]{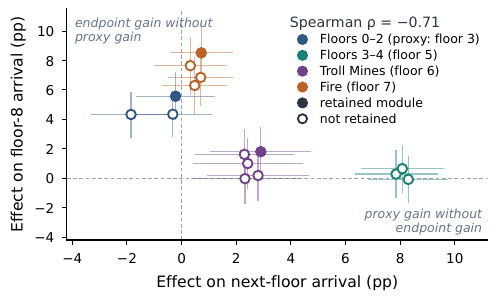}
\caption{Round-15 modules: effect on arrival at the floor that ends each region
(the local proxy) against the effect on floor-8 arrival, in percentage points
with 95\% intervals. Filled markers are retained modules.}
\label{fig:proxy}
\end{figure}

\paragraph{Local proxies and repeated use.}
We audited the 152 candidate edits of rounds 15--16 (nine of them were retained). We started with the 15 modules of round 15, and compared for each one its effect on next-floor arrival with its effect on floor-8 arrival. Both effects are negatively rank-correlated (Spearman $\rho=-0.71$, Figure~\ref{fig:proxy}), with a single module having a supported gain on both. E.g., three variants for floors 3--4 looked promising as they improve floor-5 arrival by about eight points, but we did not retain them since they have no supported gain on the final floor. Conversely, there is an early-floor variant that reduces floor-3 arrival by 1.85 points while improving floor-8 arrival by 4.33 (both intervals exclude zero). In other words, which edits look useful depends on the endpoint chosen.

Inside a goal the situation is similar. Round 16 has nineteen candidates with a favorable local measure, and not one of them has a supported floor-8 gain. One example is a lighting excursion: it is completed in one action and it reduces damage over a short window, but arrivals go down, both for a single invocation and under repeated use. In addition, fifteen candidates with a positive one-invocation estimate were also tested repeatedly, and nine of them shrink or change sign there. That is why all accepted modules were selected under repeated use from entry states, scored by the endpoint.

\paragraph{Candidate accounting.}
The ledger has 56 edits from round 15 and 96 from round 16 (152 in total), but only 24 candidates were tested both with a single invocation and under repeated use. Nineteen had a positive repeated-use screen and went to a disjoint confirmation; twelve of them kept their sign and three had intervals above zero. Recall that the ledger records an adaptive search, and thus includes superseded variants and implementation fixes as well.

\paragraph{Goal ownership.}
Inspecting the logs, we noticed that roughly a third of the episodes which get to the Ice Realm end with the agent dead in an excursion to an earlier floor. Our response in round 15 was to make the excursions a goal of their own, with its own coding agent, alongside the Ice agent and the other regions. On the common cohort, the Ice pass rate then rises from 0.41 to 0.64. Since there was no persistent runtime until round 16, we attribute this gain to the decomposition and the ownership.

\paragraph{Models and decisions.}
The world model had a specific error, as it did not predict point-blank projectile attacks; we fixed this, and added a safety filter that vetoes the actions predicted to be fatal by rollouts based on public observations. These two fixes modify the actions immediately. Their audits therefore compare trajectories that diverge, and acceptance is decided by the endpoint tests. A resource model came later. It is inert, in the sense that it only tracks the time until each resource runs out and the cost of trips. We tried a planner that reads it on every floor: the awake steps without water were halved (from 71.7 to 37.0) but floor-8 arrival dropped by 2.34 points, the reason being that the planner routes the agent past guarded water. Restricting the planner's scope helps. This is a case in which better predictions and a better local statistic do not give a better final decision.

\paragraph{Goal and model examples.}
Take for example a kit goal that raises the armor at the entry of the Troll Mines from 4.30 to 4.54 and completes 60\% of its calls. Its effect on floor-8 arrival is only $+0.01$ points, with interval $[-0.61,0.64]$ over 8,516 paired continuations. This happens because a plate can replace an iron piece instead of filling an empty
slot, and producing it can displace a diamond pickaxe. Goals that are placed on
trips the policy already makes are more effective: the strategic kit-goal
composition increases arrival from 30.91\% to 34.06\% on 4,096 reserved worlds,
which is $+3.15$ points with interval [1.86, 4.44]. Note that this cohort is
different from the one where it is placed at 31.45\% in
Table~\ref{tab:alllineage}.

In the world-model campaign we found point-blank shots that the model did not
predict. After repairing the model, the unpredicted projectile hits go down from
1,778 to 825 on a diagnostic bank of 400 continuations. The safety filter runs 24
short rollouts, eight steps each, with a model that only uses public beliefs. We
measured its effect on the endpoint together with the model repair, as one
composition, so the reset gain in Table~\ref{tab:frozen} cannot be attributed to
prediction accuracy alone. Later on, module refinement adds an inert resource
model, and the editing moves from regions of the episode to modules for
navigation, supply, combat and equipment, which span several floors. Because these
modules share state, integration tests are still needed.

\subsection{Recursive Follow-ups}
\label{app:recursive}
We report the outcomes of all five campaigns in Table~\ref{tab:recursive}. Under
the declared final comparison, only campaign 3 promotes its candidate, and its
gain of $+3.81$ points is relative to the incumbent of the campaign. If we compare
it instead with its immediate internal parent, on the same cohort of 1,024 worlds,
the final child adds only $+0.49$ points with interval $[-2.47,+3.45]$. This means
that we cannot say how much this last child contributes on its own.

\begin{table*}[t]
\centering\small
\begin{tabular}{@{}lrrrrl@{}}
\toprule
Campaign & Worlds & Before & After & Gain & Decision\\
\midrule
R1 & 4,096 & 47.63 & 48.39 & +0.76 & retain incumbent\\
R2 & 4,096 & 47.09 & 48.32 & +1.22 & retain incumbent\\
R3 & 1,024 & 44.43 & 48.24 & +3.81 & retain candidate\\
R4 & 1,024 & 49.22 & 48.34 & -0.88 & retain incumbent\\
R5 & 1,024 & 51.46 & 51.07 & -0.39 & retain incumbent\\
\bottomrule
\end{tabular}

\caption{All recursive arrival-focused campaigns. Rates are percentages and gains
are percentage points, each on the frozen world cohort of its own campaign. The
incumbent can be retained even when the point estimate of the candidate is
positive, if the promotion criterion is not met.}
\label{tab:recursive}
\end{table*}

\subsection{Boss-Victory Extension}
\label{app:boss}

If the external objective is to win the game, then arrival becomes a subgoal
itself. After freezing the Final checkpoint, we refine the policy towards the
native \texttt{defeat\_necromancer} achievement within 100,000 actions:
\begin{equation}
 J_{\mathrm{win}}(P)=\Pr_P(E_8)\Pr_P(\mathrm{win}\mid E_8).
\end{equation}
All the edits are only active on floor 8, so they do not change the states in
which the agent arrives there. This is useful, because the saved arrivals can
then be used for paired continuation comparisons and we do not need to replay
the whole approach each time. There were three campaigns, which refine the
combat, search and recovery children, and each one freezes a candidate before it
goes through a separate reset test (Table~\ref{tab:boss}).

The first candidate extends the safety filter and repairs the finishing attacks.
The second one does not pass the promotion rule, so it is not installed. The
third one is the one we retain. It adds progress
tracking based on public observations: when the search stalls it triggers
lighting and repairs the firing line, and in safe windows it lets the agent
prepare potions.

\begin{table*}[t]
\centering\small
\begin{tabular}{@{}lrrrrl@{}}
\toprule
Campaign & Worlds & Before & After & Gain & Decision\\
\midrule
R1 & 256 & 10.16 & 21.88 & +11.72 & retain candidate\\
R2 & 512 & 25.00 & 27.93 & +2.93 & retain incumbent\\
R3 & 512 & 20.70 & 25.20 & +4.49 & retain candidate\\
\bottomrule
\end{tabular}

\caption{Campaigns for boss victory, each on its own frozen cohort. Before and
after are victory percentages, and the gain is in percentage points. Campaign 2
was rejected, but we keep it in the record. Campaign 3 starts from the policy
accepted before it.}
\label{tab:boss}
\end{table*}

Then we evaluated on the common cohort of 2,048 worlds. There, the boss policy (selected in advance) wins 581/2,048 episodes (28.37\%), compared with 183/2,048 (8.94\%) before the boss refinement. In total 453 wins are gained and 55 are lost, which is a paired gain of 19.43 points with interval [17.45, 21.42]. The two policies reach floor 8 in exactly the same 1,034 worlds, and for the 1,014 episodes that do not arrive the records are identical. If we condition on arrival, the probability of winning rises from 17.70\% to 56.19\%. Note that the gain is due to the complete composition, and that this cohort is not the one used in the campaign tests.

\section{Prompt Examples}
\label{app:prompts}

This appendix reproduces some of the prompts behind the experiments of Section~\ref{sec:experiments}. There are two kinds. The first ones are the messages that we typed to the root coding agent, which are short and state an objective. The second ones are the briefs that a root agent wrote for its subagents after analysing the current policy, which are much longer (1,300 to 2,800 words) and from which we only show excerpts. All the text is verbatim, typos included, except that omissions are marked with \texttt{[...]}, absolute paths are replaced by \texttt{<repo>}, and a few symbols are written in ASCII. In these prompts, level 8 is floor 8, r16 is round 16, and HHMM was our working name for the goal hierarchy of Section~\ref{sec:framework}.

\paragraph{Regime A.}
The rounds with whole-episode feedback were started with the following message. From the moment the deeper floors were first reached, we appended the second one, changing the name of the floor.
\begin{lstlisting}[basicstyle=\scriptsize\ttfamily,breakindent=0pt,breakatwhitespace=true]
Here we have a fast implementation of the Craftax full environment. Instead of traning a deep learning model for RL'ing it, please, keep improving and optimizing the code-based policy at policy/full_policy.c (ie, the policy is a program) that maximizes the reward
\end{lstlisting}
\begin{lstlisting}[basicstyle=\scriptsize\ttfamily,breakindent=0pt,breakatwhitespace=true]
[...] As a goal, the policy should beat the final level, graveyard much more frequently (current best policy only reaches that level a handful of times, so maybe you can analyze those seeds and branch/fork from them)
\end{lstlisting}

\paragraph{Regime B.}
This is the message for the second per-floor round, after the first one had worked (a table with the results of that round is omitted).
\begin{lstlisting}[basicstyle=\scriptsize\ttfamily,breakindent=0pt,breakatwhitespace=true]
We are using coding agents (like you) to design and optimize a code-based policy that plays the craftax (full) environment. After a few runs of improvements, the policy plateaud at around 80 mean reward, with only 0.2% of seeds arriving at the final level (and stating a simple goal like "maximize arrivals at the last level). After several strategies, we found out that to improve the arrivals at the final level, the best strategy was to optimize the conditional probabilities of arriving at each level (so final is multiplicative). With this, we achieved: [...]
the new policy is at policy/full_policy.c.

Now, we want to keep exploiting that strategy (like recursively optimizing each level or interesting macro-states) to have something like 50% of arrivals at level 8. Please, analyze it (you can study the environment source code), and continue optimizing the policy. You can use subagents with gpt-6.1-sol to explore more efficiently
\end{lstlisting}

\paragraph{Regime C.}
Excerpt of the brief that the stream agents of round 14 received.
\begin{lstlisting}[basicstyle=\scriptsize\ttfamily,breakindent=0pt,breakatwhitespace=true]
## The method for this round: recursive macro-state optimization with savestates

1. **Macro-states.** The episode is a chain of macro-states, and P(8) = P(reach M) x V(M), where V(M) = P(reach floor 8 | in macro-state M). Macro-states are floor arrivals and finer phases inside a floor: cleared, cleared with the exit known, the first return trip from a floor, and so on. [...]
2. **Savestate banks.** The baseline policy was played on 16,384 development worlds (CLI seed 42000000). It saved the full engine state, future RNG and controller memory the first time each trigger fired. `bank.py resume` continues every snapshot with **your** policy. You never re-simulate the prefix, and the endpoint is about 35x more frequent than from reset.
3. **Counterfactual replicas.** Replica 0 continues with the saved dice and reproduces the original episode exactly. Replica r>0 rekeys future engine randomness, giving the same world and maps with different dice. Two policies resumed on the same bank and replicas are **paired** (common random numbers). [...]
4. **Backward induction.** Improve V on the latest macro-states first, then move earlier, always measuring the true endpoint (floor 8) from the bank. Everything downstream of the snapshot is included, so an "improvement" that sends the agent onward weaker is penalised automatically. [...]
\end{lstlisting}

\paragraph{Regime D.}
Excerpt of the brief of round 15, followed by the message that started round 16.
\begin{lstlisting}[basicstyle=\scriptsize\ttfamily,breakindent=0pt,breakatwhitespace=true]
1. **Modules and invocations.** Behaviour is a sequence of goal invocations: a floor visit, an excursion (resupply/enchant trip away from the frontier floor), a fight, a heal. Each invocation has an entry state, a termination (completed, abandoned, suspended, died, timeout) and a *continuation value*: the probability of floor 8 from the state it leaves behind. A module is good if it leaves good futures, not if it merely completes (local success .9 then .2 = .18 loses to .8 then .5 = .40).
\end{lstlisting}
\begin{lstlisting}[basicstyle=\scriptsize\ttfamily,breakindent=0pt,breakatwhitespace=true]
Can you keep working from here? (the other agent is stopped). Remember, the goal is to achieve 50% of level 8 arrivals, using our recursive optimization and the HHMM framework from our note. You can use subagents with gpt-6.1-sol to explore and work more efficiently
\end{lstlisting}

\paragraph{Regime E.}
The messages that started the world-model run, the kit-goal run and the module run, in this order.
\begin{lstlisting}[basicstyle=\scriptsize\ttfamily,breakindent=0pt,breakatwhitespace=true]
What if we:
1. refactor the current best policy (from r16, no RL tuning) so the WM is more explicit (ie a separate module) while having the same performance
2. then we can optimize policy and WM programs alternatively (but with our original framework from the paper, code-based etc))
\end{lstlisting}
\begin{lstlisting}[basicstyle=\scriptsize\ttfamily,breakindent=0pt,breakatwhitespace=true]
Have a look at our HHMM framework and harness from <repo>/framework.tex and the evolved policy we have here that reaches final floor in craftax (full) around 30% of time. Let's do a final experiment run (from that final policy at r16) to optimize the final level arrival rate, by extending the HHMM with more subgoals that can be strategic (ie not only floor levels, but also top diamond armor etc). You can use up to 3 subagents (opus-5.5) to experiment more efficiently
\end{lstlisting}
\begin{lstlisting}[basicstyle=\scriptsize\ttfamily,breakindent=0pt,breakatwhitespace=true]
What if there is a better search strategy, following our previous run with the world model and the recursive optimization theme? The idea would be to further refactor the policy(and/or WM) into submodules, and optimize each one
\end{lstlisting}

\paragraph{Control from scratch.}
The complete message of the Scratch control of Section~\ref{sec:development}, in which the framework is given as a note and there is no initial policy.
\begin{lstlisting}[basicstyle=\scriptsize\ttfamily,breakindent=0pt,breakatwhitespace=true]
Here we have a fast implementation of the Craftax full environment. Instead of training a deep learning model for RL'ing it, please, design and optimize a code-based policy (ie, the policy is a program) in policy/full_policy.c. As a goal, the policy should reach the final level, graveyard, as often as possible. Follow the framework in framework.tex. The policy must implement the interface in policy/policy_interface.h and is scored by policy/score_full.c, which you must not modify (you may extend the engine for your own tools, but the policy is scored against the unmodified engine). You can use subagents (with gpt-6.1-sol, at most 5 at a time). Do not use evaluator seeds 81000000-81004095, 82000000-82002047, 83000000-83004095 or 84000000-84002047; they are reserved for the final evaluation. Work only inside this repository. Commit every time you adopt a new best policy. At the end, freeze the best policy in benchmarks/scratch/frozen/ with its source hashes, and write a short report of what you tried.
\end{lstlisting}

\paragraph{Port and factorization.}
The two messages of the port (Section~\ref{sec:parity}), the one that started the extraction, and an excerpt of the brief for the three subagents of its second stage.
\begin{lstlisting}[basicstyle=\scriptsize\ttfamily,breakindent=0pt,breakatwhitespace=true]
Have a look at the latest code-based policy and the corresponding paper <repo>/framework.tex . For further research, our goal is to optimize some property of this policy, for instance, inference time. If you think this is an interesting experiment (or there is an even better property?), please launch an optimization run towards this
\end{lstlisting}
\begin{lstlisting}[basicstyle=\scriptsize\ttfamily,breakindent=0pt,breakatwhitespace=true]
Perfect! Now, I was wondering about optimizing for some other property. For example, is there something to gain/win of we port the C code to C++ and use some of the newest features? Be creative, and launch a new search run as a new experiment
\end{lstlisting}
\begin{lstlisting}[basicstyle=\scriptsize\ttfamily,breakindent=0pt,breakatwhitespace=true]
Thanks, interesting! Let's work on 1, please. I'm interested in the decompilation into I and p, following the Futamura projection. Specially, can we do it from the optimized C++ version (combo) at <repo>/benchmarks/policy_speed/cpp/work/combo
\end{lstlisting}
\begin{lstlisting}[basicstyle=\scriptsize\ttfamily,breakindent=0pt,breakatwhitespace=true]
[...] But most of the strategy is still C inside I: the routines the rules call as natives. **Level 2 moves the strategy into `p`, so that I keeps only game mechanics and generic algorithms.** You own one area of I. Your job is to move your area's strategy into TR without changing a single action. [...]

**Stays in I (machine).** Code a different objective would use unchanged, such as collecting achievements or killing the boss fast. [...]

**Moves to p (strategy).** Everything that decides what to do, when and where: floor-specific behaviour, phases (troll/fire/icetrip/shallow); thresholds and budgets; trips, supply schedules and goals; which skill to use, combat tactics; the shield's override policy.
\end{lstlisting}

\paragraph{Goal compilers.}
Excerpt of the prompt of the five shared searches of Section~\ref{sec:generalization}. The prompt of \texttt{Per-task} has the same structure, with one known test task in place of the training tasks and 8 tuning worlds.
\begin{lstlisting}[basicstyle=\scriptsize\ttfamily,breakindent=0pt,breakatwhitespace=true]
You are one independent gpt-6.1-sol coding-agent run in an experiment on reusable goal compilers. Work only in benchmarks/goal_composition/synthesis/run_RUN_ID (and outputs automatically created by the approved search CLI). Do not edit any existing module, harness, frozen machine, base policy, grammar, monitor, guard, or another agent's files. [...]

Optimize a self-contained Python emit(goal) module returning {'defs': TR helper definitions, 'rule': exactly one TR controller rule, 'anchor': 'rule r_strategic [build] {'}. The compiler supplies SQ_N, sq_floor[], sq_stock[], sq_hold[], sq_done(), sq_target(), sq_need(r), sq_reserve(r), and fixed progress/maintenance. The progress fields a.intent.* are read-only to your emitted controller. Base policy survival rules retain priority. You may use existing machine primitives and goal structure, but no goal-name-dependent output, seed access, result access, external I/O, changing monitor/guards, or adding runtime model calls. [...]

[...] Inspect NO validation/test goal files, task_manifest.json, other shared searches, direct searches, or held-out/validation results. [...] Do not run a custom evaluator or request additional trajectories. All new feedback must come from the three approved rounds.

Budget: exactly 3 evaluated candidates, each 16 training tasks x4 fixed worlds with50000-action horizon. [...]
\end{lstlisting}

\end{document}